\documentclass[11pt,a4paper]{article}
\usepackage[hyperref]{emnlp2020}
\usepackage{times}
\usepackage{latexsym}

\usepackage{graphicx}
\usepackage{microtype}

\aclfinalcopy 

\title{CIA\_NITT at WNUT-2020 Task 2: Classification of COVID-19 Tweets Using Pre-trained Language Models}

\author{ Yandrapati Prakash Babu \\
  Department of Computer Applications\\
NIT Trichy, India\\
  \texttt{prakash.babu23@gmail.com} \\\And
  Rajagopal Eswari \\
  Department Computer Applications\\
NIT Trichy, India\\
  \texttt{eswari@nitt.edu} \\ 
  }
\date{}

\begin{document}
\maketitle
\begin{abstract}
This paper presents our models for WNUT 2020 shared task2. The shared task2 involves identification of  COVID-19 related informative tweets. We treat this as binary text classification problem and experiment with pre-trained language models. Our first model which is based on CT-BERT achieves F1-score of 88.7\% and second model which is an ensemble of CT-BERT, RoBERTa and SVM achieves F1-score of 88.52\%.

\end{abstract}

\section{Introduction}

As of  September 07,2020 COVID-19 Coronavirus infected 27.3M people and caused 887K deaths\footnote{https://www.worldometers.info/coronavirus/}. Real time updates regarding the number of infected cases and death cases is given in dashboards. These dashboards make use of information from social networking sites like twitter. As majority of the tweets posted online are uninformative, it is necessary identify the informative tweets which include useful information related to recovered, suspected, confirmed and death cases as well as location or travel history of the cases.

The WNUT 2020 shared task2 involves identification of informative tweets. We treat this as binary text classification problem. Prior to 2018, most of the text classification models are based on Convolutional Neural Network (CNN) or Recurrent Neural Network(RNN). These models are shallow in nature and cannot learn more informative features from the input. Moreover as these models are to trained from scratch, they require more number of training instances \cite{kalyan2020bertmcn,kalyan2020secnlp}.  

Recently  pre-trained language models like BERT \cite{bert2019}, RoBERTa \cite{roberta} achieved significant improvements in many of the natural language processing tasks \cite{qiu2020pre}. BERT is a transformer encoder based language model trained using 16 GB text corpus using language modeling and next sentence prediction objectives. The 16GB text corpus includes 3.5B words from Wikipedia articles and 0.8B words from Books. BERT model is available in two versions namely BERT-base (consists of 12 transformer encoder layers with 768 hidden vector size) and BERT-large (consists of 24 transformer encoder layers with 1024 hidden vector size). As BERT models are trained using generic less noisy text corpus, these may not be effective for noisy text like tweets. Moreover,these models don't include any domain specific information. A common strategy is  to adapt BERT model to a specific domain is to further pre-train the model  or train the model from scratch using domain specific text.

In this paper, we propose two models to identify informative COVID-19 tweets. First model is based on Covid-Twitter-BERT (CT-BERT) which is a BERT-Large based model which is further trained on 160M Corona virus related tweets \cite{covid}. Second model is ensemble of CT-BERT, RoBERTa and SVM \cite{svm}. As CT-BERT is initialized from BERT-large weights and further pre-trained on COVID tweets, it has two advantages compared to BERT-large which is pre-trained on generic less noisy texts. First advantage is, CT-BERT includes domain as well as specific information and second advantage is, CT-BERT can better handle noisy texts like tweets. Our CT-BERT based model achieves F1-score of 88.87\% and ensemble  model achieves F1-score of 88.52\%

\section{Methodology}
\subsection{Dataset and Pre-Processing}

\begin{table}[h]
\begin{tabular}{p{7.5em}p{3em}p{4em}p{2em}}
\hline \textbf{Label} & \textbf{Training} & \textbf{Validation}& \textbf{Test} \\ \hline
INFORMATIVE & 3303 & 472 & 944\\
UNINFORMATIVE & 3697 &	528 &	1056\\
\hline
\end{tabular}
\caption{\label{dataset1} Original Split of Dataset}
\end{table}

\begin{table}[h]
\begin{tabular}{lcc}

\hline \textbf{Label} & \textbf{Training} & \textbf{Validation} \\ \hline
INFORMATIVE & 3024 & 751\\
UNINFORMATIVE & 3376 &	849\\
\hline
\end{tabular}
\caption{\label{dataset2}After Splitting the Dataset}
\end{table}

The dataset contains 20K tweets each of which is labeled as 0 (uninformative tweet) or 1 (informative tweet) \cite{covid19tweet}. The dataset is divided into train, validation and test sets(Actual 2k test tweets mixed with the 10K tweets), total test set size 12K. The statistics of  the original dataset is reported in Table \ref{dataset1} and the dataset splitted into 80\% and 20\% reported in Table \ref{dataset2}. 

As tweets are noisy in nature, we do the following pre-processing steps
\begin{itemize}
    \item remove unnecessary punctuation and non-ASCII characters.
    \item standardize words with repeating characters (e.g. coooool $\to$ cool)
    \item replace emoji characters with their text descriptions\footnote{We gather list of emojis and corresponding descriptions from https://emojipedia.org/}
    \item replace interjection words with their meanings (e.g. oww $\to$ pain)
    \item replace contraction with full form (e.g., I'm $\to$ I am)
    \item replace twitter slang words with related words (e.g., 2morrow $\to$ tomorrow
\end{itemize}
 
\subsection{Model Description}
We treat the problem of identification of informative tweets as binary text classification. Following the recent trend of using pre-trained language models in NLP, we propose models based on BERT and RoBERTa.
\begin{figure}[h]
    \centering
\graphicspath{ {ctbert.jpg} }
\includegraphics[width=9cm, height=11cm]{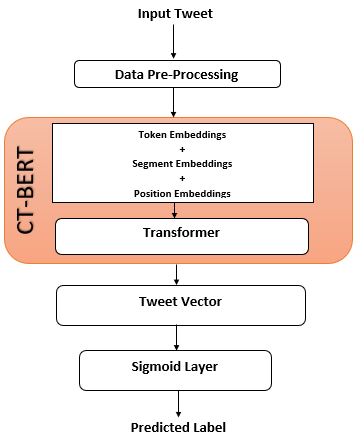}
    \caption{Overview of Model-1}
    \label{fig:ctbert}
\end{figure}

\noindent \textbf{Model-1}  This model is based on COVID-Twitter-BERT (CT-BERT). CT-BERT is initialized from BERT-Large weights and further pre-trained on 160M Corona virus related tweets. As it is binary classification, a fully connected sigmoid layer is included on the top of CT-BERT. The entire model (CT-BERT + fully connected sigmoid layer) is then fine-tuned using the training dataset. The original tweet is added with the special tokens [CLS] and [SEP] and then tokenized using word-piece tokenizer. The embedding of each token is obtained by the summation of word-piece, position and segment embeddings. A sequence of 24 transformer encoder layers is applied on these token embeddings to get the final hidden state vectors. Following , we treat $e_t \in R^h$ the final hidden vector of [CLS] token as the representation of tweet. Then, $e_t$ is passed through fully connected sigmoid layer to get the required label $\hat{p} \in [0,1]$(as shown in figure \ref{fig:ctbert}). 

\begin{equation}
    e_t = CT\-BERT (tweet)
\end{equation}

\begin{equation}
    \hat{p} = Sigmoid(W^Te_t + b)
\end{equation}

\begin{figure}[h]
    \centering
\graphicspath{ {model2.jpg} }
\includegraphics[width=8.5cm, height=11cm]{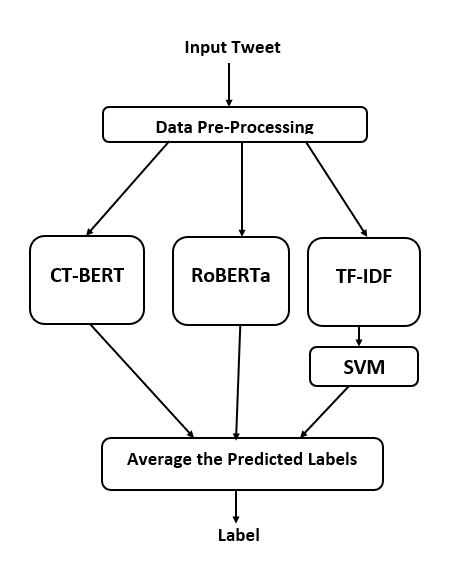}
    \caption{Overview of Model-2}
    \label{fig:model2}
\end{figure}

\noindent \textbf{Model-2} This model is  ensemble of CT-BERT, RoBERTa and TF-IDF with SVM. In this model we used base model of Roberta and TF-IDF is used for to extract the features from the tweets which were used in the SVM. Each model is individually trained using the training set. In case of CT-BERT and RoBERTa, task-specific classifier layer having fully connected sigmoid layer is added and the entire model is fine-tuned. In case of SVM, the model is trained using the tf-idf vectors of training tweets and we use kernel as sigmoid. The final prediction is obtained from the average of predictions of all these models(as shown in figure \ref{fig:model2}). 
\subsection{Evaluation Metrics}
The model is officially evaluated using precision, recall and F1-score metrics. \\

\[ Precision = \frac{T_{positive}}{T_{positive}+F_{positive}} \]
  
\[ Recall = \frac{T_{positive}}{T_{positive}+F_{negative}} \]

\[ F1-Score = 2 X \frac{Precision * Recall}{Precision + Recall} \]

\subsection{Implementation Details}
Task organizers provided  training and validation sets with labels . We merged both training and validation set and  split into 80\% train and validation sets with 80\% and 20\% of instances. We set  batch size = 32, learning rate = 3e-5 and epochs=3 after doing random search over the hyperparameter space. All our models are implemented using tranformers library in PyTorch \cite{pytorch}. 
\section{Related work}

Text classification is one of the core NLP tasks. It involves assigning labels to text sequences like phrases, sentences or documents. It has applications in various NLP tasks like sentiment analysis, spam classification, abusive text detection etc \cite{minaee2020deep}. The use of deep learning models for text classification started with using models like Convolutional Neural Network or Recurrrent Neural Network \cite{kim-2014-convolutional, nowak2017lstm}. These models are used on the top of word embeddings. To over the issue of Out of Vocabulary (OOV) words, char level CNN or RNN are used \cite{zhang2015character}. As these models are shallow in nature and need to be trained from scratch, it requires more number of training instances to train these models.  Recently, with the introduction of deep pre-trained language models like BERT, RoBERTa , there is no need to train the downstream model from scratch. To adapt the model to downstream task, it is enough to add task specific layers and fine-tune the model for few epochs \cite{bert2019,roberta}. 

\section{Results}
\begin{table}[h]
\begin{tabular}{lrrl}

\hline \textbf{Model} & \textbf{F1 Score} & \textbf{Precision}& \textbf{Recall} \\ \hline
CT-BERT & 88.87 & 87.72 & 90.04\\ \hline
CT-BERT+\\RoBERTa+\\(TFIDF+SVM) & 88.52 &	89.24 &	87.82\\

\hline
\end{tabular}
\caption{\label{results}F1-score, Precision, and Recall of proposed models on Test data}
\end{table}

To identify informative tweets related to Corona virus, we experimented with two models. First model is based on CT-BERT and second model is an ensemble of CT-BERT, RoBERTa and SVM. The results is reported in Table \ref{results}. CT-BERT based model achieved F1-score of 88.87\% and ensemble model achieved F1-score of 88.52\%. From the Table \ref{results}, it is clear that CT-BERT based model achieved slightly better results compared to ensemble model.

\section{Conclusion}
In this work, we present our models to identify  COVID-19 related informative tweets. We treat this as binary text classification problem. We propose two models based on pre-trained language models for this task. Our model based on CT-BERT achieved F1-score of 88.87\%.

\bibliographystyle{acl_natbib}
\bibliography{anthology,emnlp2020}

\end{document}